\documentclass[runningheads]{llncs}
\usepackage[T1]{fontenc}
\usepackage{graphicx,verbatim}
\usepackage{amsmath,amssymb}
\usepackage{svg}
\usepackage{booktabs}
\usepackage{multirow}
\usepackage{makecell}
\begin{document}
\title{MedPCFM-TED: One-Step Point Cloud Flow Matching for Implant Generation via Teacher-Guided Endpoint Distillation}
\titlerunning{MedPCFM-TED}
%
\author{Kamil Kwarciak\inst{1}\orcidID{0000-0002-1392-4291} \and
Marek Wodzinski\inst{1,2}\orcidID{0000-0002-8076-6246}}
\authorrunning{K. Kwarciak and M. Wodzinski}
%
\institute{Department of Measurement and Electronics, AGH University of Krakow, Krakow, Poland \and
Sano Centre for Computational Medicine, Krakow, Poland
\email{\{kwarciak,wodzinski\}@agh.edu.pl}}


\maketitle              
\begin{abstract}
Cranial implant generation is an important task in medical imaging. Recent point cloud based generative methods, particularly flow matching, offer strong reconstruction quality and efficient sampling, but still require multiple neural function evaluations during inference. This limits rapid generation of multiple plausible implant candidates. We propose \textbf{T}eacher-guided \textbf{E}ndpoint \textbf{D}istillation (\textbf{TED}), a simple one-step distillation framework for conditional cranial implant generation on point clouds. TED trains a one-step student using teacher-guided endpoint supervision and geometric matching losses, while avoiding explicit path straightening. We evaluate TED on the SkullFix and SkullBreak benchmarks. TED achieves the best overall performance on the SkullBreak dataset, remains competitive on SkullFix, and provides the strongest Chamfer distance performance among the compared one-step methods. In addition, TED generates implants in approximately 0.04s per sample. These results show that one-step distillation can substantially accelerate conditional point cloud implant generation without sacrificing reconstruction quality.


\end{abstract}

\section{Introduction}
Cranial implant design remains an important problem in medical imaging, as accurate implant reconstruction is essential for restoring skull integrity and achieving satisfactory functional and aesthetic outcomes. This task can be viewed as a 3D shape completion problem, where the goal is to reconstruct missing anatomy from incomplete skull geometry~\cite{li2023towards}. Recent approaches increasingly rely on deep learning~\cite{sulakhe2022crangan,friedrich2023point,wodzinski2022deep,wodzinski2023high,mainprize2020shape,li2023sparse}. Many of these methods operate on volumetric representations~\cite{wodzinski2022deep,mainprize2020shape,li2023sparse}, which are effective but computationally expensive and memory-intensive. More efficient alternatives include point cloud-based representations~\cite{friedrich2023point,wodzinski2023high,sulakhe2022crangan}, which are particularly attractive due to their simplicity, flexibility, and efficient processing.

Earlier methods typically approached implant reconstruction as a deterministic prediction problem, including deep-learning-based solutions~\cite{wodzinski2022deep,mainprize2020shape,li2023sparse}. More recent works, however, cast implant design as a conditional generative task on point clouds~\cite{sulakhe2022crangan,friedrich2023point,kwarciak2026medpcfm}, which is better suited to modeling the ambiguity and anatomical variability of plausible reconstructions. In this setting, diffusion models~\cite{ho2020denoising} and flow matching methods~\cite{lipman2022flow} have emerged as powerful generative frameworks. Beyond images, these model families have also been successfully applied to 3D point cloud generation~\cite{luo2021diffusion}, including conditional implant generation from defective skull geometry~\cite{friedrich2023point,kwarciak2026medpcfm}. Among them, flow matching is particularly attractive because it typically requires fewer neural function evaluations (NFEs) during sampling than diffusion models, enabling faster generation.

A related line of work aims to further reduce sampling cost by distilling generative models into one-step predictors. Progressive Distillation~\cite{salimans2022progressive} achieves this through repeated teacher-student compression, but requires several intermediate distillation stages. Methods more closely related to point cloud generation include PSF~\cite{wu2023fast} and the IMLE-based objective used in MoFlow~\cite{fu2025moflow}. PSF relies on an additional path-straightening stage prior to one-step distillation, which increases training complexity, while IMLE requires generating multiple student candidates and optimizing only the one closest to the teacher, resulting in higher computational cost and potentially less stable optimization. Thus, although existing one-step methods enable faster sampling, they often do so at the cost of more involved training.

In this work, we investigate one-step cranial implant generation using flow matching on point clouds. We introduce \textbf{T}eacher-guided \textbf{E}ndpoint \textbf{D}istillation (\textbf{TED}), a simple distillation framework in which a one-step student is trained to directly predict the final implant from the initial noise and conditioning skull geometry. TED leverages supervision from a multi-step teacher and geometric matching losses, while avoiding explicit path-straightening. Our goal is to further accelerate generation while preserving the advantages of a strong conditional generative framework. Such fast generation is particularly appealing when multiple plausible implant candidates must be generated and compared efficiently.

\section{Methods}

\begin{figure}[t]
  \centering
  \includegraphics[width=\linewidth]{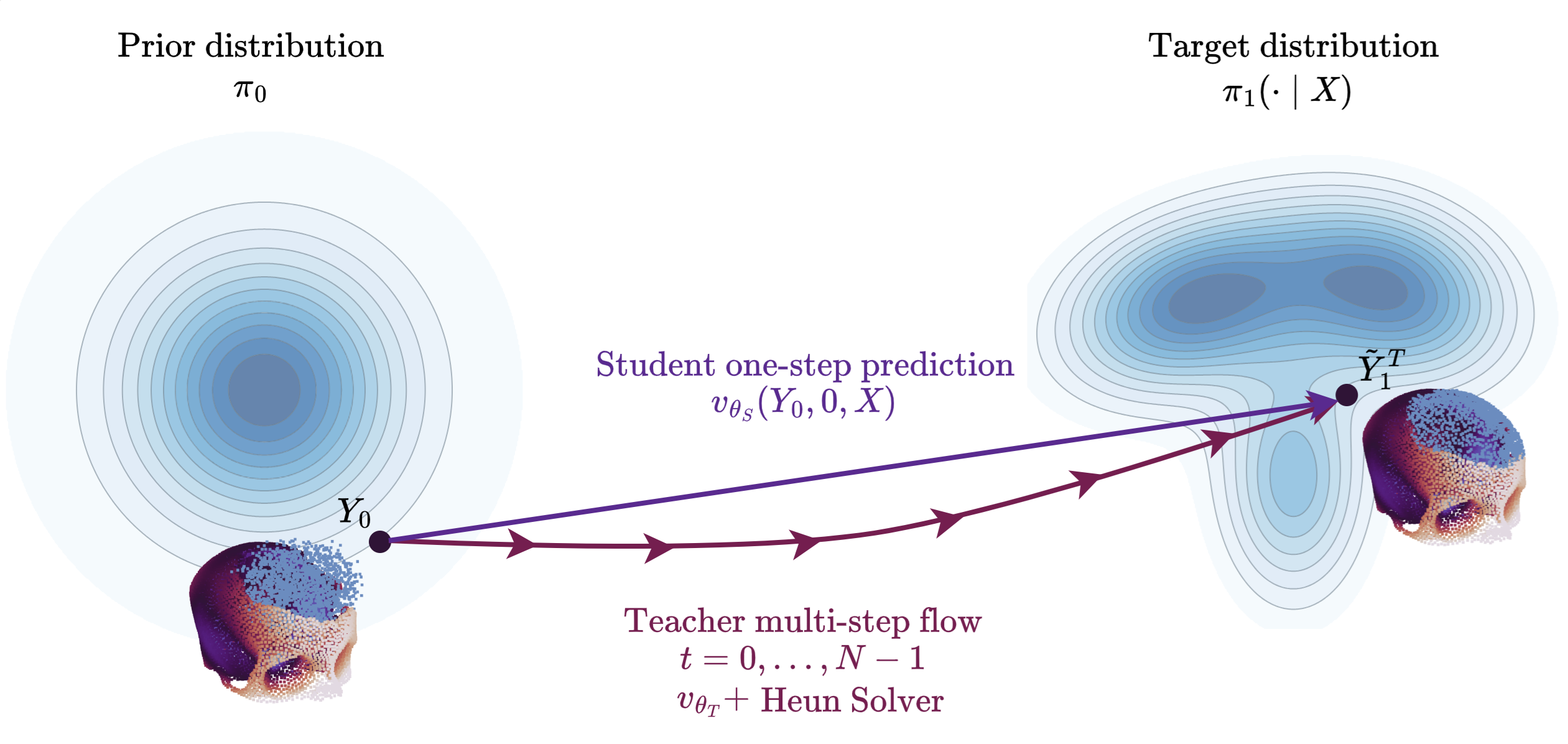}
  \caption{\textbf{Teacher-guided Endpoint Distillation (TED).} Given defective skull $X$, the flow is defined only over the implant $Y$. A prior sample $Y_0 \sim \pi_0$ is transported by the teacher to endpoint $\tilde{Y}_1^T$, while the student learns a single-step displacement from $Y_0$ to the implant prediction conditioned on $X$. Skull visualizations show conditioning context only.}
  \label{fig:scaling}
\end{figure}

We propose a simple yet effective teacher-guided distillation strategy for one-step medical point cloud flow matching, enabling a direct jump to the endpoint with a single neural function evaluation (NFE). Our method builds on the same mean-squared-error (MSE) regression objective used in prior point cloud diffusion and flow matching approaches, such as PCDiff~\cite{friedrich2023point} and PCFM~\cite{kwarciak2026medpcfm}. In addition, we adopt the Chamfer-based distillation component introduced in PSF~\cite{wu2023fast} and apply it at two levels: between the student's prediction and the teacher's prediction, and between the student's prediction and the ground-truth target. By directly constraining geometric similarity in the point cloud space, these Chamfer-based terms promote stronger shape alignment and improve the fidelity of the generated implant.

\subsection{Preliminaries}
We describe point cloud flow matching using the terminology of PCDiff~\cite{friedrich2023point} and PCFM~\cite{kwarciak2026medpcfm}. Let $X \in \mathbb{R}^{N_X \times 3}$ denote the defective skull and $Y \in \mathbb{R}^{N_Y \times 3}$ the corresponding ground-truth implant, such that the completed skull is given by $P = X \cup Y$. Our goal is to learn a conditional continuous-time flow that transforms a simple base distribution into the implant distribution conditioned on $X$. Let $Y_0 \sim \pi_0$ be a sample from an isotropic Gaussian base distribution, and let $Y_1 \sim \pi_1(\cdot \mid X)$ denote a target implant conditioned on $X$. We seek to model a time-dependent probability path that transports $Y_0$ to $Y_1$. Following the standard formulation of conditional optimal transport~\cite{lipman2024flow}, we use the affine interpolation path
\begin{equation}
    Y_t = \psi(Y_0 \mid Y_1) = tY_1 + (1-t)Y_0.
\end{equation}
The corresponding target velocity along this path is
\begin{equation}
    u_t = \frac{dY_t}{dt} = Y_1 - Y_0.
\end{equation}

Flow matching learns a conditional vector field $v_{\theta}(\cdot,t,X)$ that transports $\pi_0$ to $\pi_1(\cdot \mid X)$ through the ODE
\begin{equation}
    \frac{dY_t}{dt} = v_{\theta}(Y_t,t,X).
\end{equation}
The flow matching objective regresses the learned velocity field to the target velocity along the path~\cite{lipman2022flow}:
\begin{equation}
\mathcal{L}_{FM} =
\mathbb{E}_{t\sim \mathcal{U}[0,1],\,Y_0\sim \pi_0,\,Y_1\sim \pi_1(\cdot|X)}
\Big[
\| v_\theta(Y_t,t,X) - u_t \|_2^2
\Big].
\label{eq:FM}
\end{equation}

\subsection{Teacher-Guided Endpoint Distillation}
We now introduce our \textbf{T}eacher-guided \textbf{E}ndpoint \textbf{D}istillation (\textbf{TED}) method. We begin with a teacher flow matching model trained with the objective in Equation~\ref{eq:FM}, which provides a pretrained teacher vector field $v_{\theta_T}$.

Given an initial noise sample $Y_0 \sim \pi_0$ and the corresponding ground-truth implant $Y_1 \sim \pi_1(\cdot \mid X)$, the frozen teacher generates a target implant by integrating its conditional flow from $t=0$ to $t=1$ using an $N$-step Heun solver. We initialize the trajectory at
\begin{equation}
    Y^{(0)} = Y_0.
\end{equation}
For $n = 0, \ldots, N-2$, the teacher trajectory is computed as
\begin{equation}
\tilde{Y}^{(n+1)} =
Y^{(n)} + \frac{1}{N-1} \, v_{\theta_T}\!\left(Y^{(n)}, \frac{n}{N-1}, X\right),
\end{equation}
followed by the Heun correction step
\begin{equation}
\begin{split}
Y^{(n+1)} = Y^{(n)} + \frac{1}{2(N-1)} \Big[
v_{\theta_T}\!\left(Y^{(n)}, \frac{n}{N-1}, X\right)
\\ + v_{\theta_T}\!\left(\tilde{Y}^{(n+1)}, \frac{n+1}{N-1}, X\right)
\Big].
\label{eq:heun}
\end{split}
\end{equation}
The final teacher-generated implant is then
\begin{equation}
    \tilde{Y}_1^T = Y^{(N-1)}.
\end{equation}
We define the teacher residual as
\begin{equation}
    \Delta_T = \tilde{Y}_1^T - Y_0,
\end{equation}
which represents the total displacement induced by the teacher from the initial noise sample $Y_0$ to the final teacher-generated implant $\tilde{Y}_1^T$.

We then train a one-step student model $v_{\theta_S}$ that evaluates its velocity field only once, at $t=0$:
\begin{equation}
    \Delta_S = v_{\theta_S}(Y_0, t=0, X).
\end{equation}
The student predicts the implant as
\begin{equation}
    \hat{Y}_1 = Y_0 + \Delta_S,
\end{equation}
which corresponds to a single discretization step, similarly to PSF~\cite{wu2023fast}.

To define the TED objective, we retain the standard flow matching regression principle and match the student residual to the teacher residual using an MSE term. In addition, we introduce two geometric supervision terms based on the Chamfer distance: one between the student's prediction and the teacher-generated implant, and another between the student's prediction and the ground-truth implant. The resulting objective is
\begin{equation}
\begin{split}
    \mathcal{L}_{TED} =
    \mathbb{E}_{Y_0 \sim \pi_0,\, Y_1 \sim \pi_1(\cdot \mid X)}
    \Big[
    \lambda_{\Delta} \| \Delta_S - \Delta_T \|_2^2
    + \lambda_T \, CD(\hat{Y}_1, \tilde{Y}_1^T) \\
    + \lambda_{GT} \, CD(\hat{Y}_1, Y_1)
    \Big],
\label{eq:TED}
\end{split}
\end{equation}
where $CD(\cdot,\cdot)$ denotes the Chamfer distance. This formulation encourages the student to reproduce both the teacher's global displacement and the target implant geometry. Unlike prior one-step distillation methods that depend on trajectory straightening, such as Rectified Flow~\cite{liu2022flow} and its one-step variants~\cite{wu2023fast,liu2023instaflow}, our approach avoids explicit path-straightening. Our motivation is that, in conditional implant generation, the defective skull $X$ already provides a strong structural constraint during teacher training. This conditioning narrows the space of plausible completions and may implicitly encourage trajectories that are sufficiently direct for effective one-step distillation. The general idea of TED is presented in Figure~\ref{fig:scaling}.

\section{Experiments}

\subsection{Datasets}
We use the SkullFix and SkullBreak datasets~\cite{kodym2021skullbreak} in their combined form, following the MedPCFM setup~\cite{kwarciak2026medpcfm}. For each sample, we represent the input with 16,384 points in total, including 14,746 points sampled from the defective skull and 1,638 points sampled from the target implant. Accordingly, we set $N_X = 14{,}746$ and $N_Y = 1{,}638$. The combined train/validation/test split consists of 90/10/110 samples from SkullFix and 510/60/100 samples from SkullBreak, resulting in 600 training, 70 validation, and 210 test samples in total. All experiments are conducted on this unified benchmark.

\subsection{Experimental Setup}
We first train a PCFM model with Point Transformer V3~\cite{wu2024point} as the backbone, which serves as the teacher model for distillation. We follow the training procedure reported in PCFM~\cite{kwarciak2026medpcfm}. Specifically, all experiments are conducted on an NVIDIA GH200 GPU with 96GB of VRAM. The teacher model is trained for 15,000 epochs using the Adam optimizer~\cite{kingma2014adam} with a learning rate of $10^{-4}$, weight decay of $10^{-4}$, batch size 16, and 2,000 warm-up steps. We additionally employ exponential moving average (EMA) with a decay rate of 0.999.

We then train TED by initializing the student as a trainable copy of the teacher network. Unless stated otherwise, we use the same training hyperparameters as for the teacher model. To generate teacher implants $\tilde{Y}_1^T$, we integrate the teacher flow using the Heun ODE solver according to Equation~\ref{eq:heun}. In all TED experiments, we use $N = 40$ NFEs to generate the teacher predictions. For the TED objective in Equation~\ref{eq:TED}, we observe that all loss components have comparable magnitudes. Therefore, we set $\lambda_{\Delta} = \lambda_T = \lambda_{GT} = 1$.

For a fair comparison with alternative one-step training strategies, we also evaluate several related methods. For PSF~\cite{wu2023fast}, the procedure is more direct, as it only requires teacher-generated samples. These are obtained using the same 40-step Heun solver, and the objective is implemented by setting $\lambda_{\Delta} = 0$, $\lambda_{GT} = 0$, and $\lambda_T = 1$. For IMLE-based objective used in MoFlow~\cite{fu2025moflow}, we follow the original formulation and train the student against the teacher prediction using the Chamfer distance. For each input, we generate $m$ student candidates and optimize only the one closest to the teacher output. In our experiments, we set $m=4$. In practice, this can be implemented similarly to PSF by setting $\lambda_{\Delta} = 0$, $\lambda_{GT} = 0$, and $\lambda_T = 1$, while selecting the closest candidate for supervision. Finally, we also study a teacher-free variant in which the student receives no teacher supervision and is optimized only with respect to the ground-truth implant. In this case, for Equation~\ref{eq:TED}, we set $\lambda_{\Delta} = 0$, $\lambda_T = 0$, and $\lambda_{GT} = 1$. 

During inference, we further exploit the stochastic nature of generative completion modeling. For a given defective skull, the model can produce multiple plausible implant predictions that capture anatomically valid variability. These samples can be used either by concatenating them to obtain a denser implant point set for meshing and voxelization, or by averaging corresponding surface points to produce a smoother, more uniform representation. Overall, stochastic inference supports both efficient dense implant construction and exploration of plausible implant variability.

\begin{figure}[!b]
  \centering
  \includegraphics[width=\linewidth]{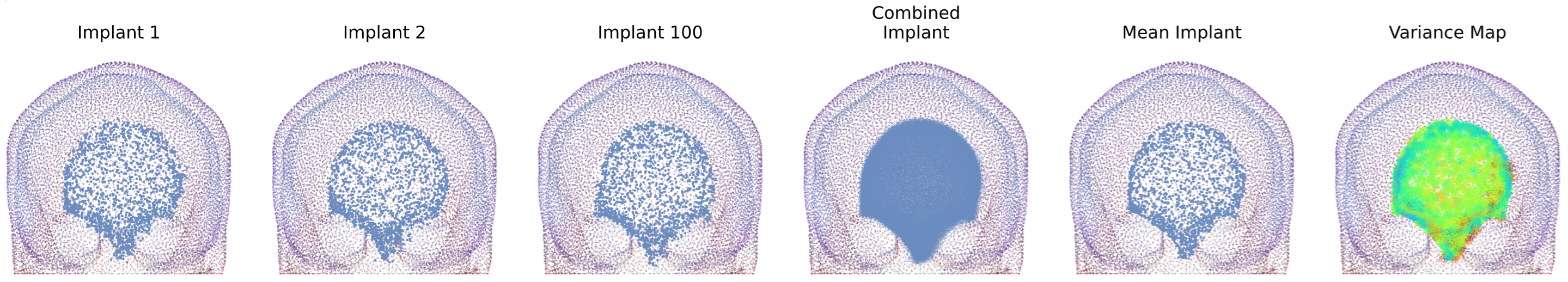}
  \caption{TED stochasticity on one SkullBreak case: three samples, mean implant, and point-wise variance map over the incomplete skull. Warmer colors indicate higher variability.}
  \label{fig:stochastic}
\end{figure}

\section{Results}
\subsection{Evaluation Protocol}
\begin{figure}[!b]
  \centering
  \includegraphics[width=\linewidth]{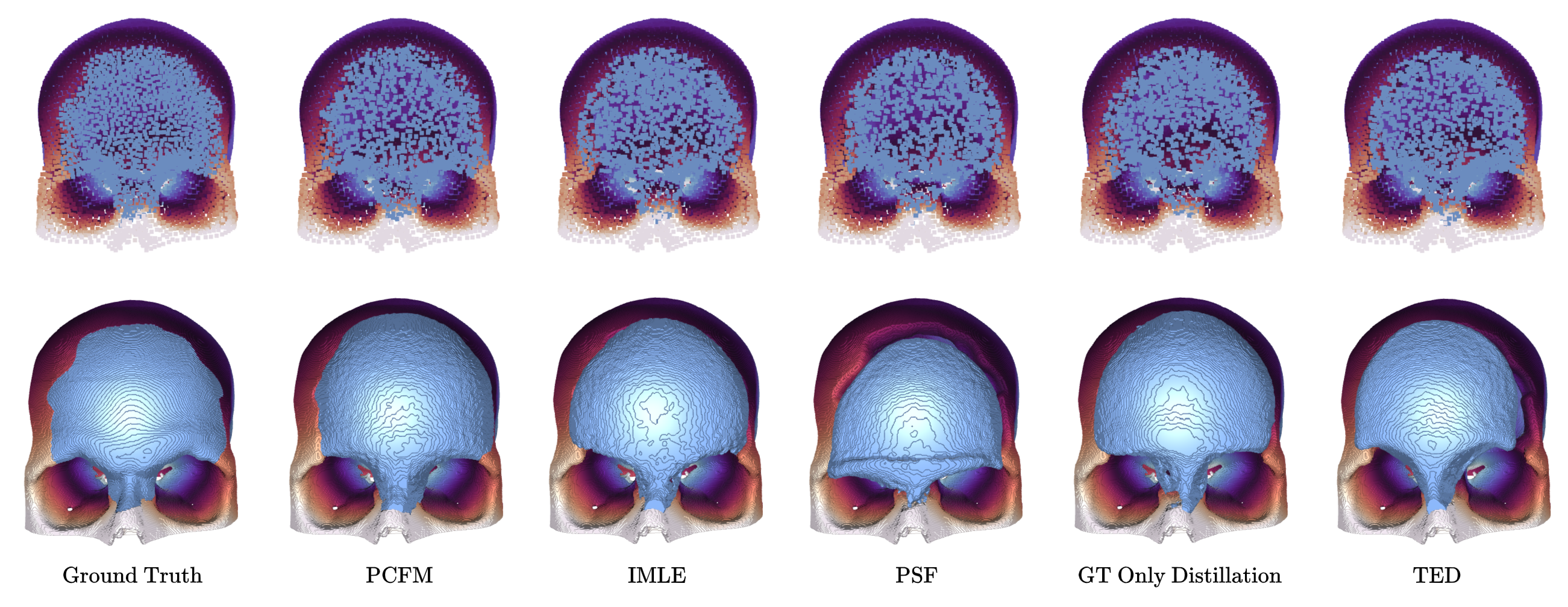}
  \caption{Qualitative SkullBreak comparison. Top: point-cloud reconstructions; bottom: voxelized counterparts. Columns show ground truth, PCFM, IMLE, PSF, GT only distillation, and TED.}
  \label{fig:skulls}
\end{figure}

We evaluate all methods on the test subsets of the SkullFix and SkullBreak datasets~\cite{kodym2021skullbreak}. As primary metrics, we report Chamfer distance and generation time. Since the considered generative models can produce multiple plausible implants for the same defective skull, Chamfer distance is computed using an aggregated mean implant from repeated stochastic generations. This mean is not obtained by averaging points with matching indices, as point cloud correspondences are not meaningful. Instead, we concatenate generated implants into a dense empirical point distribution, estimate a consensus density, and sample a fixed size representative point cloud from its high density support. This captures regions consistently predicted across stochastic samples (Figure \ref{fig:stochastic}) while remaining practical due to the fast generation speed of the distilled one-step models. To enable voxel-level comparison, we additionally report Dice similarity coefficient (DSC), boundary DSC (BDSC), and the 95th percentile Hausdorff distance (HD95). Although several point cloud-to-volume reconstruction strategies could be used, including Shape-As-Points~\cite{peng2021shape}, Poisson surface reconstruction~\cite{kazhdan2006poisson}, or occupancy-based implicit methods~\cite{peng2020convolutional}, we adopt a simple density-based voxelization procedure on the reference grid. Generated implant point clouds are aggregated into a denser point set, rasterized into a continuous voxel density map using Gaussian splatting, and binarized with automatic thresholding. Morphological closing, hole filling, and connected-component filtering are then applied to obtain the final compact implant mask. We report all results in Table~\ref{tab:skull_comparison}. We also present example completions in Figure~\ref{fig:skulls}

\newcommand{\doublemidrule}{%
  \midrule
  \addlinespace[-0.35em]
  \midrule
}

\begin{table*}[t]
\centering
\caption{Comparison on SkullFix and SkullBreak. Generation time is shared across datasets; PCFM-PTv3 with nominal one-step sampling uses two NFEs due to the Heun solver.}
\label{tab:skull_comparison}
\setlength{\tabcolsep}{4.2pt}
\renewcommand{\arraystretch}{1.18}
\resizebox{\textwidth}{!}{%
\begin{tabular}{lcc cccc cccc}
\toprule
\multirow{2}{*}{\textbf{Method}}
& \multirow{2}{*}{\makecell{\textbf{Inference}\\ \textbf{Steps}}}
& \multirow{2}{*}{\makecell{\textbf{Generation}\\ \textbf{Time [s]}}}
& \multicolumn{4}{c}{\textbf{SkullFix}}
& \multicolumn{4}{c}{\textbf{SkullBreak}} \\
\cmidrule(lr){4-7} \cmidrule(lr){8-11}
&
&
& \textbf{DSC} $\uparrow$
& \textbf{BDSC} $\uparrow$
& \textbf{HD95} $\downarrow$
& \textbf{Chamfer} $\downarrow$
& \textbf{DSC} $\uparrow$
& \textbf{BDSC} $\uparrow$
& \textbf{HD95} $\downarrow$
& \textbf{Chamfer} $\downarrow$\\
\midrule

\multirow{2}{*}{PCDiff-PTv3}
& 1    & 0.0428 & 0.028 $\pm$ 0.052 & 0.012 $\pm$ 0.034 & 39.08 $\pm$ 19.89 & 12.89 $\pm$ 1.172 & 0.002 $\pm$ 0.001 & 0.0 $\pm$ 0.0 & 35.99 $\pm$ 38.28 & 14.45 $\pm$ 0.831 \\
& 1000 & 40.899 & 0.673 $\pm$ 0.073 & 0.621 $\pm$ 0.087 & 4.414 $\pm$ 1.471 & 0.155 $\pm$ 0.048 & 0.684 $\pm$ 0.137 & 0.669 $\pm$ 0.156 & 5.001 $\pm$ 5.635 & 0.213 $\pm$ 0.132 \\
\midrule

\multirow{2}{*}{PCFM-PTv3}
& 2  & 0.0827 & 0.214 $\pm$ 0.038 & 0.083 $\pm$ 0.117 & 33.62 $\pm$ 10.13 & 0.595 $\pm$ 0.051 & 0.213 $\pm$ 0.087 & 0.107 $\pm$ 0.155 & 28.54 $\pm$ 12.98 & 0.604 $\pm$ 0.166 \\
& 40 & 3.195 & 0.824 $\pm$ 0.048 & 0.841 $\pm$ 0.046 & 3.081 $\pm$ 0.971 & 0.143 $\pm$ 0.063 & 0.744 $\pm$ 0.049 & 0.749 $\pm$ 0.056 & 3.948 $\pm$ 1.072 & 0.127 $\pm$ 0.033 \\
\doublemidrule

IMLE               & 1 & 0.0414 & 0.808 $\pm$ 0.077 & 0.789 $\pm$ 0.087 & 3.093 $\pm$ 0.915 & 0.154 $\pm$ 0.061 & 0.678 $\pm$ 0.098 & 0.630 $\pm$ 0.096 & 4.148 $\pm$ 1.168 & 0.145 $\pm$ 0.031 \\
PSF                & 1 & 0.0420 & 0.802 $\pm$ 0.067 & 0.769 $\pm$ 0.082 & \textbf{3.025} $\pm$ 0.943 & 0.157 $\pm$ 0.061 & 0.755 $\pm$ 0.083 & 0.708 $\pm$ 0.096 & 3.608 $\pm$ 1.256 & 0.152 $\pm$ 0.034 \\
TED (ours) & 1 & 0.0434 & 0.801 $\pm$ 0.070 & 0.769 $\pm$ 0.094 & 3.104 $\pm$ 1.061 & \textbf{0.149} $\pm$ 0.062 & \textbf{0.766} $\pm$ 0.079 & \textbf{0.726} $\pm$ 0.083 & \textbf{3.523} $\pm$ 1.445 & \textbf{0.144} $\pm$ 0.031 \\
GT Only (ours) & 1 & 0.0415 & \textbf{0.812} $\pm$ 0.070 & \textbf{0.808} $\pm$ 0.082 & 3.140 $\pm$ 1.054 & 0.157 $\pm$ 0.056 & 0.682 $\pm$ 0.108 & 0.626 $\pm$ 0.104 & 4.366 $\pm$ 1.392 & 0.149 $\pm$ 0.028 \\

\bottomrule
\end{tabular}%
}
\end{table*}

\subsection{Discussion}
On SkullFix, the ground truth only variant achieves the best DSC and BDSC, whereas PSF achieves the best HD95. TED, however, yields the best Chamfer distance, indicating the strongest point-level geometric fidelity among the one-step distillation techniques. This suggests that direct supervision from the ground-truth implant can be sufficient to optimize overlap-based voxel metrics on the less challenging SkullFix cases, while TED more effectively preserves fine-grained surface accuracy on the point clouds. On SkullBreak, which exhibits greater anatomical variability and more heterogeneous defects, TED consistently outperforms the remaining distillation methods. This suggests that combining teacher-guided endpoint supervision with geometric matching is more effective than relying only on teacher samples or only on ground-truth geometry, especially for more complex anatomies. The improved performance on this more challenging benchmark further indicates that TED better preserves the conditional generative behavior of the original teacher while still benefiting from explicit geometric regularization. It is also notable that both TED and PSF improve DSC over the original multi-step PCFM teacher. A plausible explanation is that the additional Chamfer-based supervision promotes better spatial coverage of the implant surface and reduces point clustering, which in turn yields more stable voxelized reconstructions after density-based conversion.

\section{Conclusion}
In this work, we introduced a simple and effective approach for one-step distillation of flow matching models for point cloud based cranial implant generation. Our method, TED, achieves the best overall performance on the SkullBreak dataset and remains competitive on SkullFix, demonstrating that fast one-step generation can be attained without sacrificing reconstruction quality. Importantly, TED enables implant generation in approximately 0.04s per sample. Given modern GPU parallelism and the advantages of stochastic completion, this makes it possible to generate around 70 implants in the time required by standard PCFM to produce a single sample, and around 940 implants in the time required by PCDiff. Such efficiency is particularly attractive for clinical implant modeling, where rapid exploration and comparison of multiple plausible implant candidates may support downstream decision-making and treatment planning.

One limitation of this work is that the voxelization procedure used for evaluation is relatively simple and could likely be improved with more advanced surface reconstruction or implicit-shape modeling techniques. However, this was not the primary focus of the study, as our main objective was to investigate one-step generative implant modeling in the point cloud domain. Moreover, voxelized representations are not strictly required in practical cranial implant workflows, where surface representations such as meshes are often more relevant.

More broadly, our results show that one-step distillation does not necessarily imply a trade-off between speed and reconstruction quality. With appropriate endpoint-level supervision, one-step generation can remain both accurate and efficient, making it a promising direction for cranial implant modeling and related conditional 3D medical generation tasks.

\begin{credits}
\subsubsection{\ackname} The project was funded by The National Centre for Research and Development, Poland under Lider Grant no: LIDER13/0038/2022 (DeepImplant). We gratefully acknowledge Polish high-performance computing infrastructure PLGrid (HPC Center: ACK Cyfronet AGH) for providing computer facilities and support within computational grant no. PLG/2026/019392. This work was partially supported by the Excellence Initiative Research University program at the AGH University of Krakow.

\subsubsection{\discintname}
The authors have no competing interests to declare that are relevant to the content of this article.
\end{credits}


%
%
%
\bibliographystyle{splncs04}
\bibliography{mybibliography}

@inproceedings{friedrich2023point,
  title={Point cloud diffusion models for automatic implant generation},
  author={Friedrich, Paul and Wolleb, Julia and Bieder, Florentin and Thieringer, Florian M and Cattin, Philippe C},
  booktitle={International conference on medical image computing and computer-assisted intervention},
  pages={112--122},
  year={2023},
  organization={Springer}
}

@article{kwarciak2026medpcfm,
  title={MedPCFM: Improving Medical Point Cloud Completion by Integrating Point Transformers and Flow Matching},
  author={Kwarciak, Kamil and Wodzinski, Marek},
  journal={arXiv preprint arXiv:2606.24433},
  year={2026}
}

@article{lipman2022flow,
  title={Flow matching for generative modeling},
  author={Lipman, Yaron and Chen, Ricky TQ and Ben-Hamu, Heli and Nickel, Maximilian and Le, Matt},
  journal={arXiv preprint arXiv:2210.02747},
  year={2022}
}

@article{lipman2024flow,
  title={Flow matching guide and code},
  author={Lipman, Yaron and Havasi, Marton and Holderrieth, Peter and Shaul, Neta and Le, Matt and Karrer, Brian and Chen, Ricky TQ and Lopez-Paz, David and Ben-Hamu, Heli and Gat, Itai},
  journal={arXiv preprint arXiv:2412.06264},
  year={2024}
}

@inproceedings{wu2023fast,
  title={Fast point cloud generation with straight flows},
  author={Wu, Lemeng and Wang, Dilin and Gong, Chengyue and Liu, Xingchao and Xiong, Yunyang and Ranjan, Rakesh and Krishnamoorthi, Raghuraman and Chandra, Vikas and Liu, Qiang},
  booktitle={Proceedings of the IEEE/CVF conference on computer vision and pattern recognition},
  pages={9445--9454},
  year={2023}
}

@article{liu2022flow,
  title={Flow straight and fast: Learning to generate and transfer data with rectified flow},
  author={Liu, Xingchao and Gong, Chengyue and Liu, Qiang},
  journal={arXiv preprint arXiv:2209.03003},
  year={2022}
}

@article{kodym2021skullbreak,
  title={SkullBreak/SkullFix--Dataset for automatic cranial implant design and a benchmark for volumetric shape learning tasks},
  author={Kodym, Old{\v{r}}ich and Li, Jianning and Pepe, Antonio and Gsaxner, Christina and Chilamkurthy, Sasank and Egger, Jan and {\v{S}}pan{\v{e}}l, Michal},
  journal={Data in Brief},
  volume={35},
  pages={106902},
  year={2021},
  publisher={Elsevier}
}

@inproceedings{wu2024point,
  title={Point transformer v3: Simpler faster stronger},
  author={Wu, Xiaoyang and Jiang, Li and Wang, Peng-Shuai and Liu, Zhijian and Liu, Xihui and Qiao, Yu and Ouyang, Wanli and He, Tong and Zhao, Hengshuang},
  booktitle={Proceedings of the IEEE/CVF conference on computer vision and pattern recognition},
  pages={4840--4851},
  year={2024}
}

@article{kingma2014adam,
  title={Adam: A method for stochastic optimization},
  author={Kingma, Diederik P and Ba, Jimmy},
  journal={arXiv preprint arXiv:1412.6980},
  year={2014}
}

@inproceedings{liu2023instaflow,
  title={Instaflow: One step is enough for high-quality diffusion-based text-to-image generation},
  author={Liu, Xingchao and Zhang, Xiwen and Ma, Jianzhu and Peng, Jian and others},
  booktitle={The Twelfth International Conference on Learning Representations},
  year={2023}
}

@article{salimans2022progressive,
  title={Progressive distillation for fast sampling of diffusion models},
  author={Salimans, Tim and Ho, Jonathan},
  journal={arXiv preprint arXiv:2202.00512},
  year={2022}
}

@inproceedings{fu2025moflow,
  title={Moflow: One-step flow matching for human trajectory forecasting via implicit maximum likelihood estimation based distillation},
  author={Fu, Yuxiang and Yan, Qi and Wang, Lele and Li, Ke and Liao, Renjie},
  booktitle={Proceedings of the Computer Vision and Pattern Recognition Conference},
  pages={17282--17293},
  year={2025}
}

@article{li2023towards,
  title={Towards clinical applicability and computational efficiency in automatic cranial implant design: An overview of the autoimplant 2021 cranial implant design challenge},
  author={Li, Jianning and Ellis, David G and Kodym, Old{\v{r}}ich and Rauschenbach, Laur{\`e}l and Rie{\ss}, Christoph and Sure, Ulrich and Wrede, Karsten H and Alvarez, Carlos M and Wodzinski, Marek and Daniol, Mateusz and others},
  journal={Medical Image Analysis},
  volume={88},
  pages={102865},
  year={2023},
  publisher={Elsevier}
}

@article{wodzinski2022deep,
  title={Deep learning-based framework for automatic cranial defect reconstruction and implant modeling},
  author={Wodzinski, Marek and Daniol, Mateusz and Socha, Miroslaw and Hemmerling, Daria and Stanuch, Maciej and Skalski, Andrzej},
  journal={Computer methods and programs in biomedicine},
  volume={226},
  pages={107173},
  year={2022},
  publisher={Elsevier}
}

@incollection{mainprize2020shape,
  title={Shape completion by U-Net: an approach to the AutoImplant MICCAI cranial implant design challenge},
  author={Mainprize, James G and Fishman, Zachary and Hardisty, Michael R},
  booktitle={Cranial implant design challenge},
  pages={65--76},
  year={2020},
  publisher={Springer}
}

@inproceedings{sulakhe2022crangan,
  title={Crangan: Adversarial point cloud reconstruction for patient-specific cranial implant design},
  author={Sulakhe, Harsh and Li, Jianning and Egger, Jan and Goyal, Poonam},
  booktitle={2022 44th Annual International Conference of the IEEE Engineering in Medicine \& Biology Society (EMBC)},
  pages={603--608},
  year={2022},
  organization={IEEE}
}

@inproceedings{luo2021diffusion,
  title={Diffusion probabilistic models for 3d point cloud generation},
  author={Luo, Shitong and Hu, Wei},
  booktitle={Proceedings of the IEEE/CVF conference on computer vision and pattern recognition},
  pages={2837--2845},
  year={2021}
}

@article{ho2020denoising,
  title={Denoising diffusion probabilistic models},
  author={Ho, Jonathan and Jain, Ajay and Abbeel, Pieter},
  journal={Advances in neural information processing systems},
  volume={33},
  pages={6840--6851},
  year={2020}
}

@inproceedings{wodzinski2023high,
  title={High-resolution cranial defect reconstruction by iterative, low-resolution, point cloud completion transformers},
  author={Wodzinski, Marek and Daniol, Mateusz and Hemmerling, Daria and Socha, Miroslaw},
  booktitle={International conference on medical image computing and computer-assisted intervention},
  pages={333--343},
  year={2023},
  organization={Springer}
}

@article{li2023sparse,
  title={Sparse convolutional neural network for high-resolution skull shape completion and shape super-resolution},
  author={Li, Jianning and Gsaxner, Christina and Pepe, Antonio and Schmalstieg, Dieter and Kleesiek, Jens and Egger, Jan},
  journal={Scientific Reports},
  volume={13},
  number={1},
  pages={20229},
  year={2023},
  publisher={Nature Publishing Group UK London}
}

@inproceedings{kazhdan2006poisson,
  title={Poisson surface reconstruction},
  author={Kazhdan, Michael and Bolitho, Matthew and Hoppe, Hugues},
  booktitle={Proceedings of the fourth Eurographics symposium on Geometry processing},
  volume={7},
  number={4},
  year={2006}
}

@article{peng2021shape,
  title={Shape as points: A differentiable poisson solver},
  author={Peng, Songyou and Jiang, Chiyu and Liao, Yiyi and Niemeyer, Michael and Pollefeys, Marc and Geiger, Andreas},
  journal={Advances in Neural Information Processing Systems},
  volume={34},
  pages={13032--13044},
  year={2021}
}

@inproceedings{peng2020convolutional,
  title={Convolutional occupancy networks},
  author={Peng, Songyou and Niemeyer, Michael and Mescheder, Lars and Pollefeys, Marc and Geiger, Andreas},
  booktitle={european conference on computer vision},
  pages={523--540},
  year={2020},
  organization={Springer}
}
\end{document}